\documentclass{article}
\usepackage{amssymb}
\usepackage{amsmath,graphicx,mlspconf}
\usepackage{xcolor}

\title{DFA-CON: A Contrastive Learning Approach for Detecting Copyright Infringement in DeepFake Art}
\name{Haroon Wahab$^{\star}$ \qquad Hassan Ugail$^{\star}$ \qquad Irfan Mehmood$^{\dagger}$}

\address{
    $^{\star}$School of Computer Science, AI and Electronics, University of Bradford, UK \\
    $^{\dagger}$School of Management, University of Bradford, UK
}

\begin{document}

\maketitle

\begin{abstract}
Recent proliferation of generative AI tools for visual content creation—particularly in the context of visual artworks—has raised serious concerns about copyright infringement and forgery. The large-scale datasets used to train these models often contain a mixture of copyrighted and non-copyrighted artworks. Given the tendency of generative models to memorize training patterns, they are susceptible to varying degrees of copyright violation. Building on the recently proposed DeepfakeArt Challenge benchmark, this work introduces DFA-CON, a contrastive learning framework designed to detect copyright-infringing or forged AI-generated art. DFA-CON learns a discriminative representation space, posing affinity among original artworks and their forged counterparts within a contrastive learning framework. The model is trained across multiple attack types, including inpainting, style transfer, adversarial perturbation, and cutmix. Evaluation results demonstrate robust detection performance across most attack types, outperforming recent pretrained foundation models. Code and model checkpoints will be released publicly upon acceptance.

\end{abstract}

\section{Introduction}
\label{sec:intro}
The growing availability of generative AI tools for visual content creation has raised critical concerns around copyright infringement, especially in the domain of visual artworks \cite{aiart-impact}. Generative models trained on large-scale, web-scraped datasets often absorb patterns from both copyrighted and public domain images, making them prone to reproducing unauthorized content \cite{van2021memorization}. This phenomenon is particularly problematic in the context of AI-generated art, where stylistic and structural similarities to original artworks may constitute legal or ethical violations \cite{somepalli2023diffusion}. Given the ease with which forged or derivative artworks can be created and distributed, there is a compelling need for automated methods to assess the originality and attribution of generative outputs.

To study and benchmark the detection of AI-generated art forgeries, the recently proposed DeepfakeArt Challenge \cite{aboutalebi2023deepfakeart} provides a comprehensive dataset comprising over 32,000 image pairs spanning a variety of generative manipulation techniques. Each entry in the dataset consists of a pair of images—either a forged/generated version of an original artwork or two dissimilar, unrelated images. The manipulated images cover several attack types including inpainting, style transfer, adversarial perturbation, and cutmix, simulating realistic scenarios of content misuse. This dataset enables the development of algorithms that go beyond pixel-level artifact detection, focusing instead on semantic similarity and visual attribution, a crucial capability when detecting copyright violations in generative art.

To address this problem, we propose DFA-CON, a supervised contrastive learning framework designed to detect copyright-infringing or forged art generated by AI models. Rather than treating forgery detection as a traditional classification problem, DFA-CON learns an embedding space that encodes semantic similarity between original and manipulated images. Using supervised contrastive loss, the model is trained to pose affinity among original artworks and their forged counterparts, while distancing unrelated images within the same batch. This representation-level formulation allows for greater generalization across manipulation types and better discrimination of subtle, high-quality forgeries, which are common in the domain of AI-generated art. 

Our main contributions are summarized as follows:
\begin{itemize}
    \item We propose a supervised contrastive learning framework, DFA-CON, for training visual encoders to detect copyright infringement in DeepFake art. To the best of our knowledge, this is the first work that introduces a dedicated model tailored specifically for infringement detection in AI-generated artworks.
    
    \item We conduct a comprehensive comparison between DFA-CON and recent general-purpose vision foundation models to assess the need for domain-specific training in the context of generative art forensics.
    
    \item We release our model checkpoints and code to facilitate future research. The codebase is modular and supports integration of any embedding model via a standardized wrapper interface, enabling rapid experimentation and reproducibility.
\end{itemize}

\section{Related Works}
\label{sec:rw}
\subsection{AI Generated Art}
\label{subsec:rw1}
Recent advancements in generative models such as GANs and diffusion models have enabled the creation of highly realistic and stylistically rich visual artworks \cite{elgammal2017can, rombach2022high}. While this has opened new avenues in creative expression, it has also raised significant concerns regarding authorship, originality, and copyright infringement \cite{somepalli2023diffusion}. Prior work in AI-generated art has largely focused on generation techniques, artistic style transfer, and aesthetics modeling, with less attention paid to post-generation verification or attribution. The DeepfakeArt Challenge benchmark \cite{aboutalebi2023deepfakeart} was introduced to bridge this gap by facilitating research on forgery and contamination detection in generative art settings.

\subsection{Deep Fake Detection}
\label{subsec:rw2}
Deep Fake detection has emerged as a critical subfield in computer vision, primarily aimed at identifying synthetic or manipulated facial content \cite{yan2023deepfakebench}. Detection methods typically focus on spatial or frequency domain inconsistencies, fine-grained artifact analysis, or classification of pixel-level distortions. However, these methods are often limited to human faces and fail to generalize to non-photorealistic domains like art. In contrast, the task of detecting forgeries in AI-generated artworks involves semantic similarity and contextual interpretation rather than artifact spotting, necessitating representation-based methods like ours.

\subsection{Contrastive Learning}
\label{subsec:rw3}
Contrastive learning has shown great promise in learning discriminative and generalizable visual representations \cite{le2020contrastive}. In particular, supervised contrastive loss \cite{khosla2020supervised} has been used effectively in domains where fine-grained similarity modeling is essential. Unlike self-supervised contrastive learning, which relies on augmentations of the same image, supervised variants utilize structured label information to align semantically similar samples. Our method builds on this paradigm by using original-forged image pairs as positives and treating dissimilar pairs as implicit negatives, resulting in an embedding space optimized for semantic attribution and forgery detection.

\section{Preliminaries}
\label{sec:prel}

\subsection{Copyright Infringement in Art}
 
\label{subsec:prel1}
Deep Fake art generative models are vulnerable to violating copyright terms by producing images that mimic or closely resemble content protected under copyright \cite{kriplani2025solidmark}. A formal mathematical formulation of copyright infringement in this context is introduced in~\cite{aboutalebi2023deepfakeart}. For clarity and contextual relevance, we present a simplified version here:

\begin{equation}
    \| A(y)_\Omega - A(T(\hat{x}))_\Omega \| < f(|\Omega|) \cdot \delta
    \label{eq:infringement}
\end{equation}

Here, $y$ denotes the generated image and $\hat{x}$ is a potentially copyright-protected training image. The operator $T(\cdot)$ represents a geometric transformation (e.g., resizing, flipping, rotation), and $\Omega$ is a region of significant size within the image. The function $A(\cdot)$ defines the domain of representation—either raw pixel space or an edge-based representation. The term $f(|\Omega|)$ is a monotonic function adjusting sensitivity based on region size, and $\delta$ is a fixed similarity threshold. Infringement is said to occur if the distance in the representation space for any region $\Omega$ falls below this threshold.
\subsection{Deepfake Art Challenge}
\label{subsec:prel2}
The DeepfakeArt Challenge~\cite{aboutalebi2023deepfakeart} provides a large-scale benchmark for detecting copyright-infringing or adversarially manipulated images in the domain of AI-generated art. The dataset comprises over 32,000 image pairs, each labeled as either similar (indicating a generated version of an original artwork) or dissimilar (completely unrelated pairs). Each similar pair corresponds to a specific form of generation-based manipulation, known as \emph{attack type}. The challenge enables the development and evaluation of models capable of distinguishing original artworks from manipulated or forged counterparts.

\subsubsection{Inpainting}
Inpainting attacks simulate localized editing by removing a region of the original image and replacing it using a generative inpainting model such as Stable Diffusion \cite{rombach2022high}. The masked region is filled in a semantically consistent but potentially infringing way.

\subsubsection{Style Transfer}
Style transfer-based forgeries are generated by transferring the artistic style of the original artwork onto a different content image, producing a stylized image that may still share strong visual resemblance to the original \cite{styletransfer}.

\subsubsection{Adversarial Perturbation}
Adversarial examples are crafted by introducing imperceptible pixel-level changes to the original image that lead to semantic drift or visual mimicry.

\subsubsection{CutMix}
CutMix attacks generate forged samples by cutting and pasting regions from multiple images, including the original. This composite manipulation results in a hybrid image that may retain recognizable features from copyrighted content.

\begin{figure*}[t]
  \centering
   \vspace{-8.5em}
  \includegraphics[width=\textwidth]{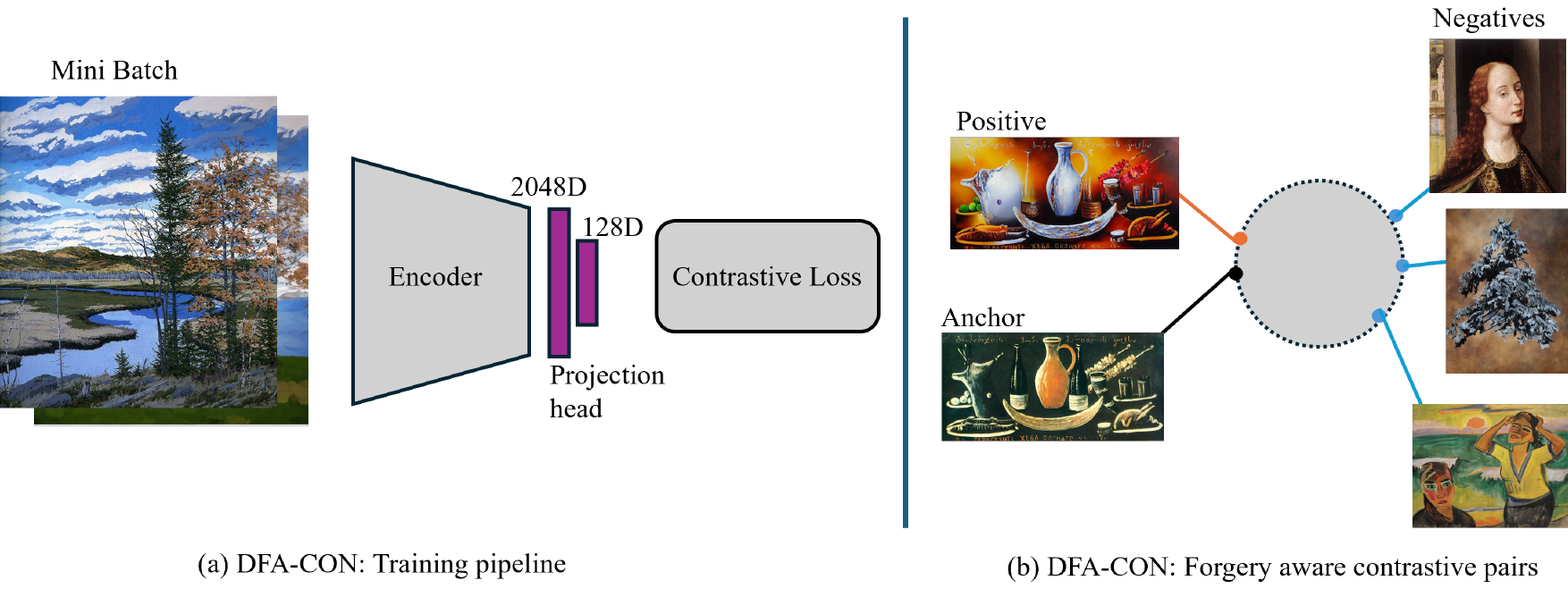}
   \vspace{-8.5em}
  \caption{Methodology for training contrastive model to detect copyright infringements in DeepFake Art (DFA-CON)}
  \label{fig:meth}

\end{figure*}

\begin{table}[htbp]
\caption{Distribution of Similar Pairs by Attack Type}
\centering
\begin{tabular}{lcc}
\hline
\textbf{Attack Type} & \textbf{\#Pairs} & \textbf{Percentage} \\
\hline
Inpainting           & 5063   & 39\% \\
Style Transfer       & 3074   & 24\% \\
Adversarial          & 2730   & 21\% \\
CutMix               & 2000   & 16\% \\
\hline
\textbf{Total}       & 12,867   & 100\% \\
\hline
\end{tabular}
\label{tab:attack-distribution}
\end{table}

\section{Methodology}
\label{sec:meth}
 This section describes the overall methodology employed in this work. We first present DFA-CON, a contrastive representation learning framework designed to detect copyright infringement in AI-generated art (see  Fig.~\ref{fig:meth}). We then introduce an inference-time detection pipeline that leverages the pretrained embedding model to evaluate whether a generated image constitutes a potential infringement (see  Fig.~\ref{fig:inf}).
 
\subsection{DFA-CON}
\label{subsec:dfa}
DFA-CON consists of three core components: (a) forgery-aware contrastive sampling, (b) representation learning using an encoder and projection head, and (c) a supervised contrastive loss objective for training.

\subsubsection{Forgery-Aware Contrastive Sampling}
\label{subsubsec:dfa-1}
In order to adapt the contrastive learning paradigm to the task of copyright-infringement detection in generative art, a sampling strategy is employed that reflects the structural assumptions of the problem as formalized in Section~\ref{subsec:prel1}. Each original artwork in the dataset is treated as an anchor instance, denoted by \( i \), and its corresponding forgeries—produced via different attack types—are collected into a set of positives \( \mathcal{P}(i) \). These forged versions represent multiple semantically similar views of the anchor image under the notion of copyright violation. A batch \( \mathcal{B} \) is formed by sampling multiple such anchors and their associated positives. For each anchor \( i \), the negative set \( \mathcal{N}(i) \) is implicitly defined as all other instances in the batch that do not belong to \( \{i\} \cup \mathcal{P}(i) \), i.e., \( \mathcal{N}(i) = \mathcal{B} \setminus (\{i\} \cup \mathcal{P}(i)) \). This formulation enables batch-wise supervised contrastive training where similarity is learned under the forgery-aware relational structure of the data.

\subsubsection{Representation Learning}
\label{subsubsec:dfa-2}

The proposed framework is structured to learn semantically meaningful representations that support contrastive learning objectives. A two-stage architecture is employed, consisting of a backbone encoder and a projection head. The encoder maps input images from pixel space into a high-dimensional semantic representation space, while the projection head transforms these representations into a lower-dimensional space optimized for contrastive loss. In this work, a ResNet-50 architecture is used as the encoder, where the final fully connected classification layer is removed. This results in a representation vector in \( \mathbb{R}^{2048} \) for each input image.

Two variants of the projection head are explored as part of an ablation study. The first variant is a linear projection head, which maps the encoder output from \( \mathbb{R}^{2048} \) directly to a 128-dimensional space. The second variant is a multilayer perceptron (MLP) that includes a hidden layer of size 2048 with a non-linear activation, followed by a projection to \( \mathbb{R}^{128} \). These representations are then used for supervised contrastive learning.

\subsubsection{Contrastive Loss}
\label{subsubsec:dfa-3}
To train the model to learn robust and discriminative representations for forgery detection, we adopt the supervised contrastive (SupCon) loss~\cite{khosla2020supervised}. SupCon is selected due to its improved robustness and its native support for the multi-positive setting, which aligns well with our data formulation, where each anchor may have multiple forged variants. By leveraging all valid positive associations for a given anchor within a batch, the loss encourages consistent representation of similar images and separation from unrelated ones.

Given a batch \( \mathcal{B} \) of size \( N \), and a representation \( \mathbf{z}_i \) for each sample \( i \), the SupCon loss for an anchor \( i \) is defined as:

\begin{equation}
\mathcal{L}_i = \frac{-1}{|\mathcal{P}(i)|} \sum_{p \in \mathcal{P}(i)} \log \frac{\exp(\mathbf{z}_i \cdot \mathbf{z}_p / \tau)}{\sum_{a \in \mathcal{B} \setminus \{i\}} \exp(\mathbf{z}_i \cdot \mathbf{z}_a / \tau)}
\label{eq:supcon}
\end{equation}

Here, \( \mathcal{P}(i) \) denotes the set of positives for anchor \( i \), and \( \tau \) is a temperature scaling parameter. The final loss is computed by averaging \( \mathcal{L}_i \) over all anchors in the batch. This formulation allows the model to leverage all known positive associations in a batch while contrasting against a shared set of negatives.

\subsubsection{Training details}
\label{subsubsec:trn}

The model is trained using the SupCon loss on mini-batches constructed with the forgery-aware sampling strategy described earlier. An 80-20\% split is applied on the official training split of the dataset ensuring no overlap between original as well as forged versions across the split. All images are resized to \(224 \times 224\) and normalized using ImageNet statistics. The encoder is initialized with ImageNet-pretrained weights, and the entire network is optimized using stochastic gradient descent (SGD) with momentum 0.9. We train the model for 50 epochs, using an initial warm-up phase of 10 epochs followed by cosine annealing of the learning rate. The base learning rate is set to 0.01. Early stopping is applied with a patience of 10 epochs based on validation loss. The temperature parameter \( \tau \) is set to 0.07, following the original SupCon formulation. Training is conducted using a batch size of 128; we also experiment with 32 and 64 as hyperparameters. Both linear and MLP-based projection heads are evaluated as part of an ablation study. 

\subsection{Copyright Infringement Detection Pipeline}
\label{subsec:detect}
The proposed inference pipeline is designed for practical copyright verification of AI-generated artworks. Consider a scenario where a user possesses a single image or a collection of original, copyright-protected artworks. The goal is to determine whether a given image produced by a generative AI model infringes upon any of the known originals.

\begin{figure}[t]
  \centering
  \includegraphics[width=\linewidth]{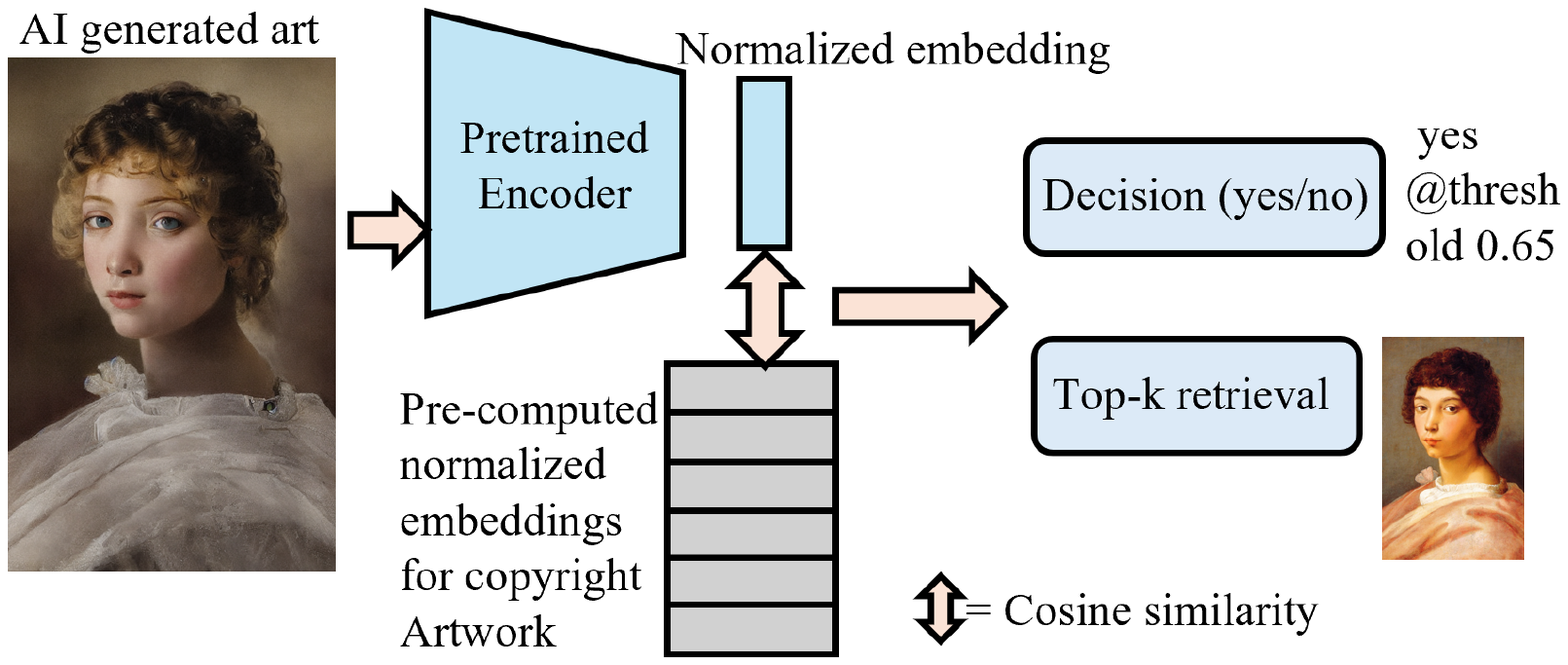}
  \vspace{-3.5em}
  \caption{Copyright infringement detection pipeline}
  \label{fig:inf}
\end{figure}

As illustrated in Fig.~\ref{fig:inf}, the generated image is first passed through the embedding model, which could be DFA-CON or any other pretrained visual encoder. The resulting embedding is normalized and compared against the set of pre-computed, normalized embeddings of the original artworks using cosine similarity. A threshold-based decision rule is then applied to determine whether the image constitutes a potential infringement. This pipeline is computationally lightweight and scalable, and it also enables top-$k$ retrieval of the most similar original artworks for use in more fine-grained analysis or human-in-the-loop verification.

\begin{table}[htbp]
\caption{Overall Performance Comparison (Precision, Recall, F1)}
\centering
\begin{tabular}{lccc}
\hline
\textbf{Model} & \textbf{P} & \textbf{R} & \textbf{F} \\
\hline
ResNet-50 (ImageNet)       & 0.7988 & 0.7330 & 0.7645 \\
ViT-B/16 (ImageNet)        & 0.813 & 0.7032 & 0.7541 \\
DINO-v2 ViT-L/14    & 0.7181 & 0.6125 & 0.6611 \\
CLIP ViT-B/16 (OpenAI)     & 0.8643 & 7056 & 0.7769 \\
\hline
\textbf{DFA-CON (Ours)}    & \textbf{0.9481} & \textbf{0.7465} & \textbf{0.8353} \\
\hline
\end{tabular}
\label{tab:overall-performance}
\end{table}

\section{Evaluation}
\label{sec:eval}

We evaluate our model on an exclusive test split provided by the DeepfakeArt benchmark. A similarity threshold is first determined using validation set and then applied during testing to make binary decisions. If the cosine similarity between the pair in inference exceeds the threshold, the model classifies the pair as similar—indicating a potential copyright violation. Otherwise, the image is considered dissimilar, suggesting no infringement. This formulation naturally aligns with a binary classification setting, and we adopt precision, recall, and F1 score as our primary performance metrics. We report both overall performance across all attack types and per-attack-type results to provide insight into the model's robustness against specific forms of forgery.

\subsection{Baseline Models}
\label{ssec:bm}
To assess the effectiveness of DFA-CON, we compare it against several widely used pretrained visual foundation models. To the best of our knowledge, there are no existing models specifically trained for the task of detecting copyright infringement in AI-generated art. Therefore, we consider recent high-capacity embedding models that are commonly used for general-purpose visual representation learning.

The baseline models include: (i) ResNet-50 pretrained on ImageNet \cite{he2016deep}, (ii) ViT-B/16 pretrained on ImageNet \cite{dosovitskiy2020image}, (iii) DINO-v2 ViT-L/14 \cite{oquab2023dinov2}, and (iv) CLIP ViT-B/16 \cite{radford2021learning}. All baseline models are evaluated using the same inference pipeline described in Section~\ref{subsec:detect}, where cosine similarity is computed between normalized embeddings and a fixed threshold is used for binary classification.

\begin{table}[htbp]
\caption{Performance by Attack Type (Precision, Recall, F1)}
\centering
\begin{tabular}{llccc}
\hline
\textbf{Attack} & \textbf{Model} & \textbf{P} & \textbf{R} & \textbf{F} \\
\hline
Inpainting     & ResNet-50      & 0.7378 & 0.9771 & 0.8407 \\
               & ViT-B/16       & 0.7634 & 0.9459 & 0.8449 \\
               & DINO-v2        & 0.6616 & 0.6405 & 0.6509 \\
               & CLIP ViT-B/16  & 0.8137 & 0.8837 & 0.8473 \\
               & \textbf{DFA-CON} & \textbf{0.9393} & \textbf{0.9386} & \textbf{0.939} \\
\hline
Style Transfer & ResNet-50      & 0.4955 & 0.4179 & 0.4534\\
               & ViT-B/16       & 0.5 & 0.4058 & 0.448 \\
               & DINO-v2        & 0.4566 & 0.4954 & 0.4752 \\
               & CLIP ViT-B/16  & 0.6233 & 0.4802 & 0.5425 \\
               & \textbf{DFA-CON} & \textbf{0.8923} & \textbf{0.9696} & \textbf{0.9294} \\
\hline
Adversarial    & ResNet-50      & 0.6864 & 0.9954 & 0.8125 \\
               & ViT-B/16       & 0.7178 & 1.0 & 0.8357 \\
               & DINO-v2        & 0.6614 & 1.0 & 0.7962 \\
               & CLIP ViT-B/16  & 0.75 & 0.9977 & 0.8563 \\
               & \textbf{DFA-CON} & \textbf{0.9168} & \textbf{0.9943} & \textbf{0.9539} \\
\hline
CutMix         & ResNet-50      & 0.5906 & \textbf{0.3623} & \textbf{0.4491} \\
               & ViT-B/16       & 0.5626 & 0.2859 & 0.3791 \\
               & DINO-v2        & 0.5601 & 0.2697 & 0.3641 \\
               & CLIP ViT-B/16  & \textbf{0.6347} & 0.3299 & 0.4341 \\
               & \textbf{DFA-CON} & 0.5341 & 0.0544 & 0.0987 \\
\hline
\end{tabular}
\label{tab:attack-type-performance}
\end{table}

\subsection{Performance Comparison}
\label{subsec: pc}
\subsubsection{Overall Performance}
As shown in Table~\ref{tab:overall-performance}, DFA-CON significantly outperforms all baseline foundation models across precision, recall, and F1 score on the overall test set. This suggests that pretrained vision models—despite being trained on large-scale and diverse datasets—do not produce task-aligned representations sufficient for detecting copyright violations in DeepFake art. In contrast, DFA-CON benefits from its supervised contrastive training on explicitly structured forgery data, enabling it to learn more discriminative and attribution-aware embeddings.

The observed performance gap highlights an important limitation of general-purpose visual encoders. While such models are effective at broad semantic understanding, they may not capture the fine-grained visual cues and structural similarities that distinguish authentic artworks from forged variants. This is especially relevant in the context of DeepFake art, where visual mimicry often occurs at a stylistic or compositional level rather than through overt image artifacts. By training on forgery-aware pairings, DFA-CON is able to better internalize the nuanced characteristics of copyright infringement in generative content.

\subsubsection{Per-Attack-Type Performance}

Table~\ref{tab:attack-type-performance} presents a breakdown of performance across individual attack types. DFA-CON consistently achieves the highest F1 scores on inpainting, style transfer, and adversarial attacks, confirming its ability to generalize across varied forgery strategies. However, performance notably declines on the CutMix attack, where DFA-CON underperforms compared to all baseline models. This unexpected result may be attributed to the fundamentally different nature of CutMix-based forgeries, which involve compositional splicing of image regions from multiple sources rather than style imitation or localized perturbations. Such hybrid manipulations may introduce ambiguous visual signals that current contrastive supervision struggles to capture effectively. While this remains speculative, it suggests that additional investigation is needed to better understand model behavior on this attack type and to explore whether tailored contrastive sampling or auxiliary supervision could improve performance in such scenarios.

\section{Ablation Study}
\label{sec:abl}
We conduct an ablation study to examine the impact of the probe point within DFA-CON, specifically evaluating whether representations extracted from different levels of the model affect detection performance. Results indicate that using embeddings directly from the encoder output in \( \mathbb{R}^{2048} \) yields the highest scores across all metrics. In comparison, probing from the projection head—whether using a linear or MLP variant projecting to \( \mathbb{R}^{128} \)—leads to a slight degradation in performance, typically in the range of 1–2\%. This suggests that while the projection head is effective during training for optimizing the contrastive loss, the encoder-level features are better aligned with the downstream task of infringement detection.

\section{Conclusion}
\label{sec:con}

This work presented DFA-CON, a supervised contrastive learning framework for detecting copyright infringement in AI-generated art. Our method leverages forgery-aware sampling and contrastive representation learning to distinguish original artworks from their forged counterparts. Extensive experiments on the DeepfakeArt benchmark demonstrate that DFA-CON significantly outperforms several widely used foundation models. We also analyzed performance across different attack types, highlighting the robustness and limitations of our approach. Ablation results showed that encoder-level representations are most effective for the task. We hope our publicly released code and model will serve as a foundation for future work in generative content forensics.

\bibliographystyle{IEEEbib}
\bibliography{refs}

\end{document}